\documentclass[letterpaper]{article} 
\usepackage{times}  
\usepackage{helvet}  
\usepackage{courier}  
\usepackage[hyphens]{url}  
\usepackage{graphicx} 
\usepackage{natbib}  
\usepackage{caption} 
\usepackage{algorithm}
\usepackage{algorithmic}

\usepackage{newfloat}
\usepackage{listings}
\usepackage{amssymb}
\usepackage{booktabs}
\usepackage{multirow}
\usepackage{subfigure}
\usepackage{amsmath}

\usepackage{xspace}

\newcommand{\ourmethod}{FedStar\xspace}
\usepackage{aaai23}  
\usepackage{times}  
\usepackage{helvet}  
\usepackage{courier}  
\usepackage[hyphens]{url}  
\usepackage{graphicx} 
\usepackage{natbib}  
\usepackage{caption} 
\usepackage{algorithm}
\usepackage{algorithmic}

\usepackage{newfloat}
\usepackage{listings}
\DeclareCaptionStyle{ruled}{labelfont=normalfont,labelsep=colon,strut=off} 
\floatstyle{ruled}
\newfloat{listing}{tb}{lst}{}
\floatname{listing}{Listing}
\title{Federated Learning on Non-IID Graphs via Structural Knowledge Sharing}
\author {
    Yue Tan\textsuperscript{\rm 1}\equalcontrib, 
    Yixin Liu\textsuperscript{\rm 2}\equalcontrib, 
    Guodong Long\textsuperscript{\rm 1}, 
    Jing Jiang\textsuperscript{\rm 1}, 
    Qinghua Lu\textsuperscript{\rm 3}, 
    Chengqi Zhang\textsuperscript{\rm 1}
}
\affiliations {
    \textsuperscript{\rm 1} University of Technology Sydney, 
    \textsuperscript{\rm 2} Monash University, 
    \textsuperscript{\rm 3} Data61, CSIRO\\
    yue.tan@student.uts.edu.au, yixin.liu@monash.edu, \{guodong.long, jing.jiang\}@uts.edu.au, \\
    qinghua.lu@data61.csiro.au, chengqi.zhang@uts.edu.au
}

\usepackage{bibentry}

\begin{document}

\maketitle

\begin{abstract}
Graph neural networks (GNNs) have shown their superiority in modeling graph data. Owing to the advantages of federated learning, federated graph learning (FGL) enables clients to train strong GNN models in a distributed manner without sharing their private data. A core challenge in federated systems is the non-IID problem, which also widely exists in real-world graph data. For example, local data of clients may come from diverse datasets or even domains, \textit{e.g.}, social networks and molecules, increasing the difficulty for FGL methods to capture commonly shared knowledge and learn a generalized encoder. From real-world graph datasets, we observe that some structural properties are shared by various domains, presenting great potential for sharing structural knowledge in FGL. Inspired by this, we propose FedStar, an FGL framework that extracts and shares the common underlying structure information for inter-graph federated learning tasks. To explicitly extract the structure information rather than encoding them along with the node features, we define structure embeddings and encode them with an independent structure encoder. Then, the structure encoder is shared across clients while the feature-based knowledge is learned in a personalized way, making FedStar capable of capturing more structure-based domain-invariant information and avoiding feature misalignment issues. We perform extensive experiments over both cross-dataset and cross-domain non-IID FGL settings, demonstrating the superiority of FedStar. 
\end{abstract}

\section{Introduction} \label{sec:intro}
Graph neural networks (GNNs) have been widely used to model graph-structured data in a variety of scenarios and applications, such as recommender systems~\cite{wu2020graph}, drug discovery~\cite{gaudelet2021utilizing}, and traffic~\cite{bai2020adaptive}. 
Most existing GNNs follow a centrally training principle where graph data need to be collected together before training~\cite{kipf2017semi,xu2019powerful}. However, nowadays, a large number of graph data are generated from edge devices and may contain private data of users, hindering the traditional GNNs from training strong models by collective intelligence~\cite{zhang2021federated}. 

Federated learning (FL), as a new machine learning paradigm, allows clients to collaboratively train a globally shared model or personalized models in a decentralized manner while not sharing the local data of  clients~\cite{yang2019federated}. 
Due to the advantages of FL, it is natural to apply FL to graph data to mitigate the data isolation and protect the security of graph data owned by end users, such as molecule graphs owned by pharmaceutical companies and social networks located in the social app of end users. 
Collaboratively training GNN models with an FL framework, 
federated graph learning (FGL) has emerged to be a promising direction to further explore the potential of GNNs on decentralized graph data.

\begin{figure}[t!]
	\centering
	\setlength{\abovecaptionskip}{-0.2cm}
	\setlength{\belowcaptionskip}{-0.6cm}
    \begin{center}
    \subfigure[]{
    	\begin{minipage}[b]{0.202\textwidth}
    		\includegraphics[width=1\textwidth]{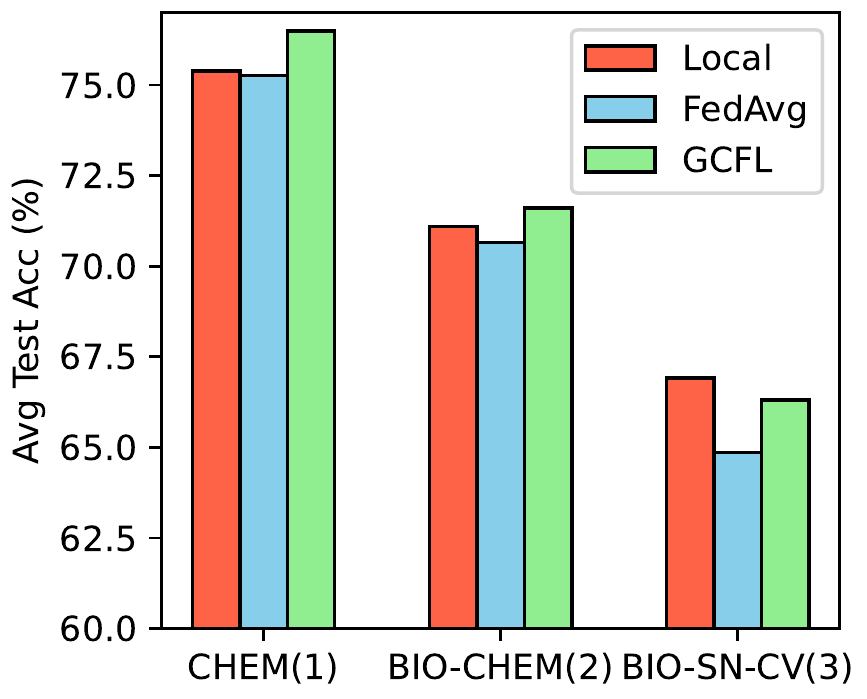}
    	\end{minipage}
    	\label{fig:intro_deter}
    }
    \subfigure[]{
    	\begin{minipage}[b]{0.238\textwidth}
    		\includegraphics[width=1\textwidth]{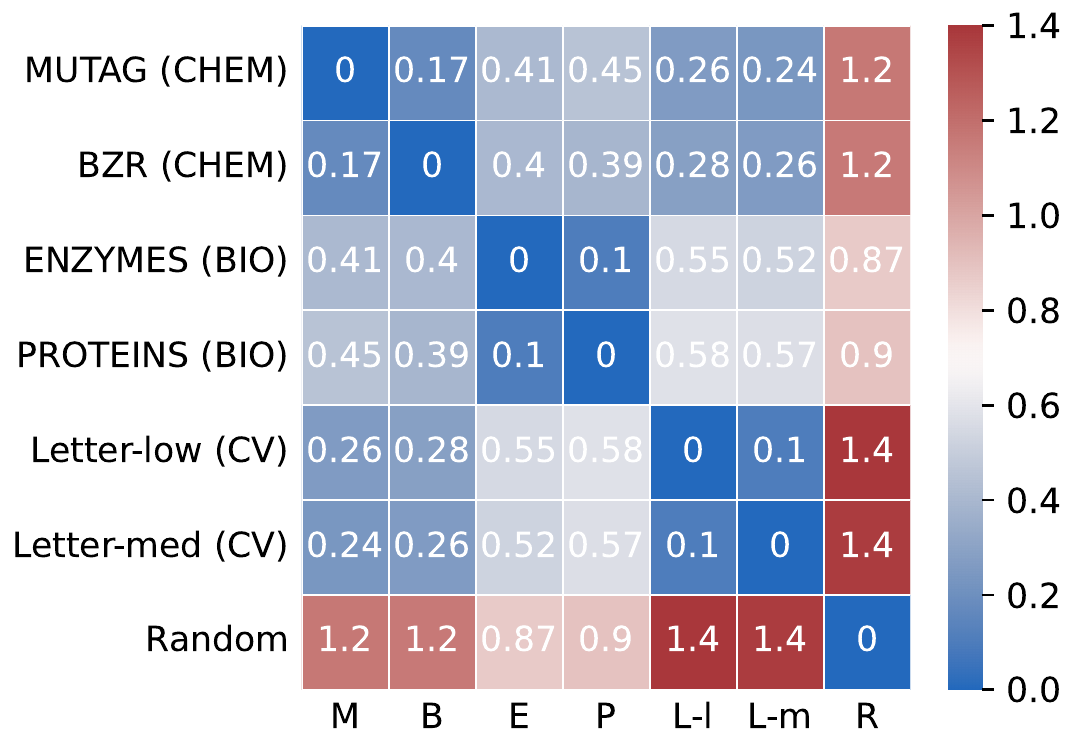}
    	\end{minipage}
    	\label{fig:intro_heatmap}
    }
    \end{center}
    \caption{(a)~Performance comparison of local training and FL/FGL methods~\cite{mcmahan2016communication,xie2021federated} in three cross-dataset/domain non-IID settings. (b)~The heatmap of Jensen-Shannon divergence of degree distributions among six real-world graph datasets from different domains and random graphs. The random graphs are generated with Erdős–Rényi model~\cite{erdos1960evolution}.} 
    \label{fig:intuitive}
\end{figure}

An important challenge in FL is the non-IID data problem where local datasets of clients can be extremely different due to their diverse behaviors or preferences. 
This difference may lead to unstable training and performance drop in some extreme cases, \textit{e.g.}, the data of clients are from different domains~\cite{li2020fedbn,kairouz2021advances}. 
The non-IID data problem, unfortunately, also naturally exists in FGL owing to the diversity of graph data. 
For example, different companies/platforms maintain rich social networks data, which are of great heterogeneity (\textit{i.e.}, diverse feature spaces and connection rules) due to different data collection manners and purposes. 
It is beneficial to overcome the non-IID issue, enabling these companies to collaboratively provide better online social network service without sacrificing data security. 
Furthermore, it is of great potential to explore the underlying knowledge commonly shared by heterogeneous graph domains, \textit{e.g.}, social networks and molecules.

An emerging branch of studies has attempted to alleviate the influence of non-IID graphs~\cite{xie2021federated,wang2020graphfl}. 
They take conventional GNNs (\textit{e.g.}, GraphSAGE~\cite{hamilton2017inductive} and GIN~\cite{xu2019powerful}) as local encoders and introduce various techniques to local training or global aggregation, such as clustering and meta-learning. 
Based on the conventional GNNs, existing FGL methods still generate representations via a feature-based message passing scheme, deteriorated by the \textit{feature heterogeneity} (\textit{i.e.}, different feature spaces and dimensions) across graphs from multiple domains. 
In FGL procedures, aggregating GNN models trained upon diverse feature spaces may fail to capture useful underlying knowledge shared over non-IID graphs, which in turn unexpectedly hurts the personalized performance of local models. 
As shown in Fig.~\ref{fig:intro_deter}, in non-IID FGL settings, the performance of federated methods, \textit{i.e.}, FedAvg and GCFL, is sometimes even worse than local training. 

Although feature information tends to be heterogeneous across different graph domains, some structural properties underlying graphs can be universally shared across different domains ~\cite{qiu2020gcc,leskovec2005graphs,xie2021federated}, inspiring us to exploit the potential of sharing structural knowledge with FGL. 
As shown in Fig.~\ref{fig:intro_heatmap}, compared with random graphs, real-world graphs share more inherent structural properties even if they are from different domains. 
Nevertheless, it is difficult for existing FGL frameworks to capture and communicate such structural knowledge. 
Firstly, these methods usually depend on conventional GNNs where the structure information is implicitly encoded into feature-based node representations; thus, heterogeneous features inevitably affect the quality of structural knowledge learning. 
Moreover, most FGL methods still perform parameter aggregation over all learnable parameters, making it hard to extract and share structure-specific knowledge across clients. 
Hence, a natural research question to ask is ``\textit{how to insulate structure information from the heterogeneous features during local training and how to perform federated knowledge sharing?}''

To answer the aforementioned question, 
in this work, we propose \textbf{Fed}erated graph learning via \textbf{St}ructure knowledge sh\textbf{ar}ing (\textbf{FedStar} for short), a framework that takes full advantage of natural structure information in graphs to deal with inter-graph FL tasks. 
Our theme is to capture and share the universal structural knowledge across multiple clients to boost the local performance of clients in FGL. 
Specifically, we first introduce a vectorial structure embedding that explicitly represents the inherent structural properties shared across different domains. Then, to avoid interference by feature heterogeneity, we design a feature-structure decoupled GNN to capture high-level structural knowledge with an independent channel while learning attributive knowledge with another one. 
Finally, we design an FL framework with structural knowledge sharing, where domain-agnostic structural information is shared across different clients and feature-based representations are learned locally. As a result, clients are capable of learning generalized structure encoders as well as personalized feature encoders simultaneously. 
The contributions of our work are three-fold: 
\begin{itemize}
    \item We study the non-IID problem in federated graph learning with tremendous heterogeneous graph data from a new perspective, \textit{i.e.}, structural knowledge sharing, which sheds good light on the future research in this field. 
    \item We propose a novel federated graph learning framework, namely FedStar, enabling clients to learn domain-invariant structural knowledge globally while capturing domain-specific feature-based knowledge locally. 
    \item We conduct extensive experiments on four cross-dataset/domain non-IID settings and show that FedStar consistently outperforms baselines by a large margin.
\end{itemize}

\section{Related Work} \label{sec:rw}
\paragraph{Graph Neural Network.}
Graph neural networks (GNNs) are a family of neural models for modeling graph-structured data~\cite{wu2020comprehensive}. Most GNNs follow a message passing scheme, where node representation is learned by aggregating and transforming the embeddings of its neighbors and itself~\cite{hamilton2017inductive,xu2019powerful}. For instance, GCN~\cite{kipf2017semi} aggregates messages via averaging the neighboring representations, and GIN~\cite{xu2019powerful} leverages a summation function for aggregation. 
Thanks to such a message passing scheme, GNNs can achieve excellent performance when learning from a single graph~\cite{velivckovic2018graph,liu2022towards} or a set of graphs with shared feature space~\cite{hu2020strategies,hu2020gpt}. However, the feature-based message initialization mechanism makes the feature space and learnable parameters in GNNs tightly coupled, leading to the difficulty of training GNNs on graph data in multiple domains~\cite{qiu2020gcc,zhu2021transfer,liu2022good}. 
Besides, a branch of GNNs extracts structural/positional encoding to explicitly represent the structure information~\cite{li2020distance,dwivedi2021graph}, inspiring us to capture universal structural patterns across different domains by GNNs. 

\paragraph{Federated Learning.}
Federated learning (FL) has attracted much attention recently owing to its potential to enable collaborative training while protecting data privacy~\cite{kairouz2021advances,yang2019federated,lyu2022privacy}. 
The standard FL algorithm, namely FedAvg, iteratively conducts local training at clients and global parameter averaging at central server~\cite{mcmahan2016communication}. 
A number of methods are proposed to improve FedAvg in terms of communication efficiency~\cite{hamer2020fedboost}, generalization ability~\cite{yuan2021we,qu2022generalized}, robustness to heterogeneity~\cite{wang2020tackling,tan2021fedproto,chen2022calfat}, etc. 
One of the existing challenges in real-world FL is the data heterogeneity, also known as the non-IID problem, where clients may have diverse label and/or feature distributions due to their various behaviours and habits~\cite{luo2021no,tan2022federated,chen2022practical}.
To tackle this, clustering-based FL methods propose to divide clients according their similarity~\cite{ghosh2020efficient,long2022multi,ma2022convergence}. 
Multiple works leverage meta-learning to improve the personalized ability of the local model~\cite{fallah2020personalized,jiang2019improving} or
introduce model decoupling scheme to enable better personalization~\cite{chen2022on}. 

\paragraph{Federated Graph Learning.}
Federated graph learning (FGL) supports distributed GNN training, extending its original application scenarios. So far, existing FGL studies can be categorized into three types, \textit{i.e.}, inter-graph, intra-graph, and graph-structured FGL~\cite{zhang2021federated}. 
In inter-graph FGL, each client owns a set of graphs and participates in FL training to learn a better GNN to model local data~\cite{xie2021federated}, learn a generalizable model~\cite{zhu2022federated}, or to model spatial-temporal graph data~\cite{jiang2022federated,lou2021stfl}. 
In intra-graph FGL, instead of complete graphs, each client only owns a subgraph of the entire graph and the learning scheme is to deal with the missing links~\cite{chen2021fede}, \textit{i.e.}, generating missing neighbors~\cite{zhang2021subgraph}, community discover~\cite{baek2022personalized}. Intra-graph FGL can be applied to financial crimes detection~\cite{suzumura2019towards}, 
In graph-structured FGL, graphs are used to model the inherent relationship among clients and can be applied to various data types, such as images~\cite{chen2022personalized} and traffic data~\cite{meng2021cross}. 
In this paper, we consider inter-graph FGL and focus on learning a better local model for each client rather than training a global GNN model.

\section{Preliminaries} \label{sec:prel}

\subsection{Graph Neural Networks} \label{sec:gnn}
Let $G=(V, E)$ be a graph consisting of a set of nodes $V$ and a set of edges $E$ connecting these nodes. Each node $v \in V$ has a feature vector {$\mathbf{x}_v$}. Based on the graph structure and node features, GNNs can be used to learn the node-level representation vector $\mathbf{h}_v$ of node $v \in V$ and/or the graph-level representation vector $\mathbf{h}_G$ of graph $G$. Existing GNNs usually follow the message passing scheme where $\mathbf{h}_v$ is iteratively updated by aggregating the representations of node $v$'s neighbors. Formally, for {an $L$-layer GNN, its $l$-th layer} can be formulated as
\begin{equation}
\mathbf{a}_{v}^{{(l)}}=\mathrm{AGGREGATE}^{{(l)}}\left(\left\{{\mathbf{h}_{u}^{(l-1)}}: u \in \mathcal{N}(v)\right\}\right),
\end{equation}
\begin{equation}
\mathbf{h}_{v}^{{(l)}}=\mathrm{UPDATE}^{{(l)}}\left(\mathbf{h}_{v}^{(l-1)}, \mathbf{a}_{v}^{{(l)}}\right),
\end{equation}
\noindent where {$\mathbf{h}_{v}^{(l)}$ is the representation vector of node $v$ output by the $l$-th layer}, $\mathcal{N}(v)$ is the set of node $v$'s neighbors. Different $\mathrm{AGGREGATE}$ and $\mathrm{UPDATE}$ strategies are adopted in different GNN variants and sometimes can be integrated together~\cite{kipf2017semi,hamilton2017inductive,gilmer2017neural,xu2018representation}. 

Specifically, for graph classification, the graph representation $\mathbf{h}_G$ can be further obtained by {aggregating all the node representations involved in graph $G$ via various graph-level readout functions, such as summation and mean pooling. }

\subsection{Federated Learning} \label{sec:fl}
In a vanilla FL setting with $M$ clients, the $m$-th client owns a private dataset $\mathcal{D}_m$. The global objective of the FL framework is 
\begin{equation} \label{eq:fl}
\min _{\left(w_{1},w_{2}, \cdots, w_{M}\right)} \frac{1}{M} \sum_{m=1}^{M} \frac{|\mathcal{D}_m|}{N} \mathcal{L}_m\left(w_{m} ; \mathcal{D}_{m}\right),
\end{equation}
\noindent where $N$ is the total number of instances over all clients, $\mathcal{L}_m$ and $w_m$ are the loss function and model parameters of client $m$, respectively. 

Standard FL methods aim to learn a globally shared model $w=w_1=w_2=\cdots=w_M$. The representative method is FedAvg~\cite{mcmahan2016communication} which periodically aggregates the model parameters of all clients at the server by
\begin{equation} \label{eq:para_agg}
\overline{w} \leftarrow \sum_{m=1}^{M} \frac{\left|\mathcal{D}_{m}\right|}{N} {w}_{m}
\end{equation}
\noindent and return the averaged model back to clients. However, this kind of methods may have poor performance due to non-identical distributions across clients. Recent studies address this problem by applying personalized techniques to local training and/or global aggregation procedure, which allows $w_m$ to perform better on the local data of client $m$.

In a federated graph learning (FGL) framework, $w_m$ and $\mathcal{D}_{m}$ refer to the parameter set of GNN model and the graph dataset at client $m$, respectively. Since we focus on classification tasks over non-IID graph datasets in this paper, \textit{e.g.}, protein function prediction, the loss function $\mathcal{L}_m$ in Eq.~\ref{eq:fl} is the commonly used cross-entropy loss.

\begin{figure*}[t!]
	\centering
	\setlength{\abovecaptionskip}{-0.0cm}
	\setlength{\belowcaptionskip}{-0.1cm}
    \begin{center}
    \subfigure[Feature-structure decoupled GNN]{
    		\includegraphics[height=4.5cm]{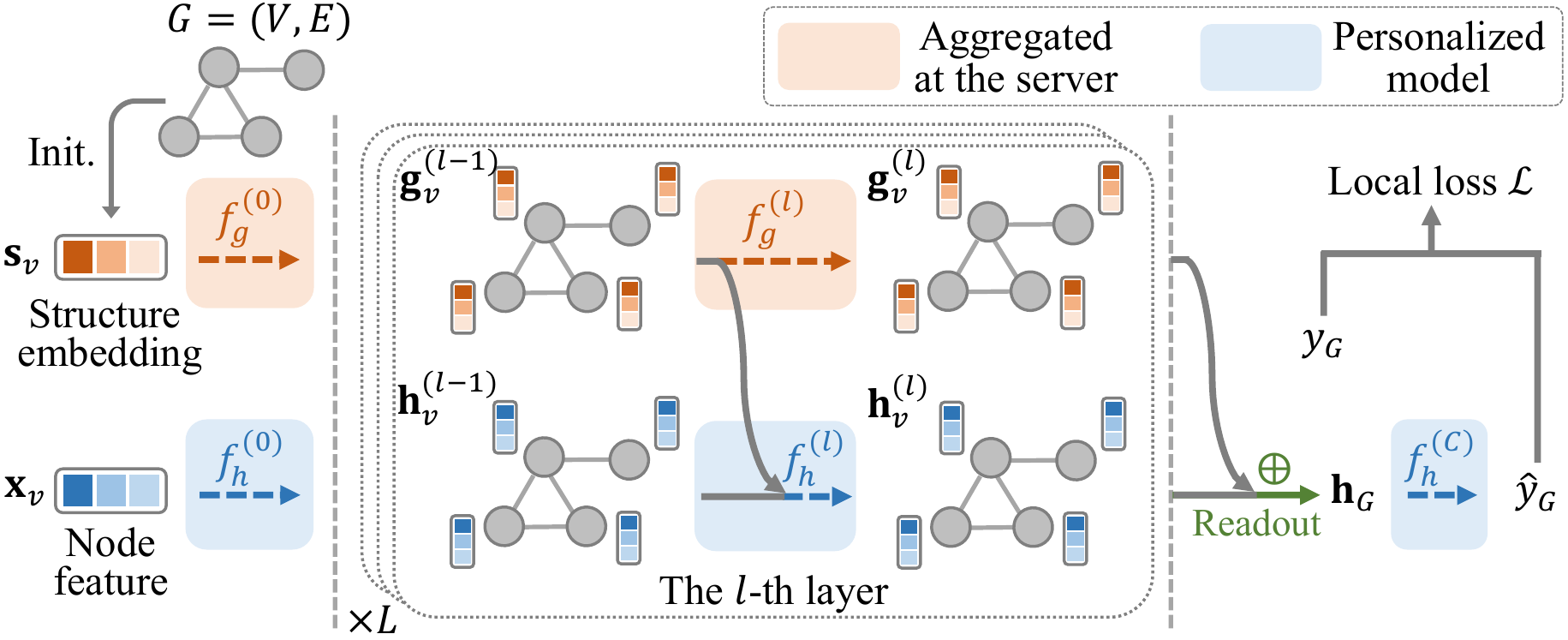}
    	\label{fig:local_gnn}
    }	
    \hspace{0.4cm}
    \subfigure[Model aggregation scheme]{
    		\includegraphics[height=4.5cm]{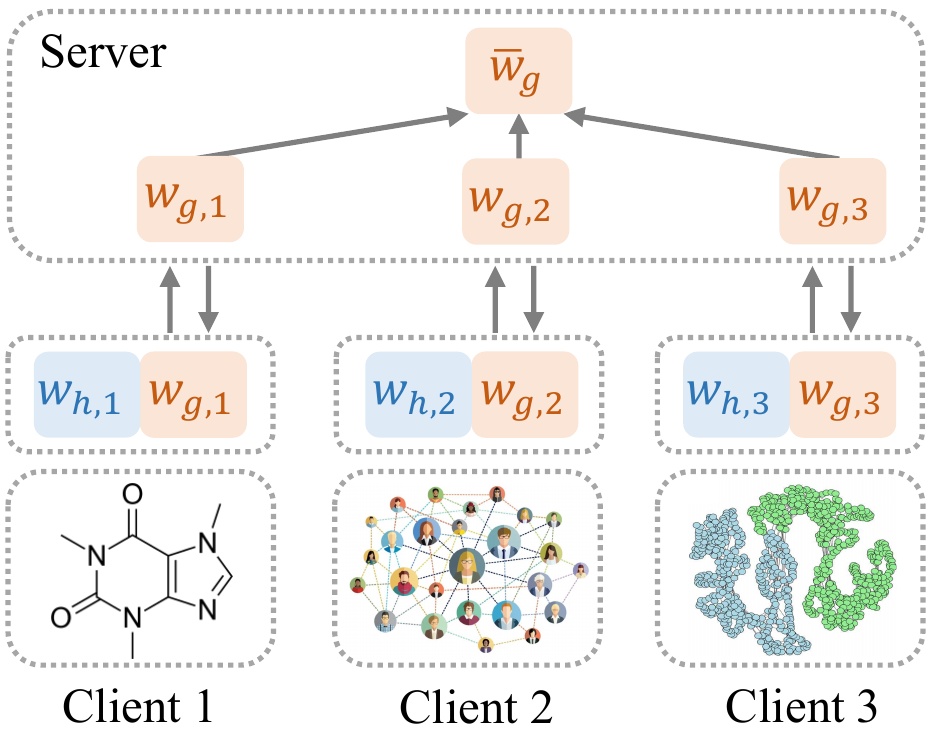}
    	\label{fig:agg_scheme}
    }
    \end{center}
    \caption{(a) An overview of the local GNN architecture in our proposed \ourmethod. Blue boxes correspond to the models that are trained locally, and personalized for each client. Orange boxes correspond to the models that are aggregated at the server, with knowledge shared across clients. (b) An illustration of the data distribution and model aggregation scheme. }
    \label{fig:framework}
\end{figure*}

\section{Methodology} \label{sec:method}
In this section, we present our \ourmethod framework in detail. 
Given a set of clients that own various graph datasets in the federated graph learning (FGL) framework, our goal is to enable each client to achieve higher performance on their own dataset by exploring universally shared underlying knowledge with other clients. 
To achieve the above goal by sharing structural knowledge across clients, we need to answer the following three questions:
\begin{itemize}
    \item How to model structure information and capture universal structural patterns across different domains? 
    \item How to design the GNN architecture to encode the feature and structure information, respectively?
    \item How to conduct structural knowledge sharing during the federated learning stage?
\end{itemize}

To answer the above questions, we first provide the definition of structure embedding and how it is initialized in our proposed method. Next, we illustrate the proposed feature-structure decoupled GNN architecture that learns attributive and structural knowledge with two channels, respectively. Finally, we discuss how to share structural knowledge with our proposed \ourmethod framework. 

\subsection{Structure Embedding Initialization}

In most conventional GNNs, the structure information is \textit{implicitly} encoded into node representations along with feature information through the procedure of feature aggregation. However, in non-IID FGL scenarios where features are usually heterogeneous across domains, 
it is difficult to uniformly encode general structure information (\textit{e.g.}, node degree) into feature-based representations from different spaces. 
To bridge the gap, we introduce a novel type of node-level information carrier, namely \textbf{structure embedding}, to \textit{explicitly} represent the universal structural knowledge in a vectorial form. The structure embedding can be initialized from the graph structure, and plays the role of ``features'' during message aggregation. In this case, the structure information is learned independently to feature information. 

To capture general and comprehensive structure information, in \ourmethod, we construct structure embedding that incorporates both local and global structural patterns. 
In particular, to capture local structural knowledge, we introduce a degree-based structure embedding (DSE) which uses vertex degrees in the form of one-hot encoding. For node $v$ in a graph, its DSE is denoted as 
\begin{equation}
\mathbf{s}_{v}^{\mathrm{DSE}}=\left[\mathbb{I}(d_v=1), \mathbb{I}(d_v=2), \cdots, \mathbb{I}(d_v \geq k_{1})\right] \in \mathbb{R} ^{\mathrm{k_1}},
\end{equation}

\noindent where $d_v$ is the degree of node $v$, $\mathbb{I}$ is the identity function, and $k_1$ is the dimension of DSE. The benefit of using DSE to express local structure information is three-fold. First, degree is a basic geometric property that can be extracted from graphs of any domain. Second, the distributions of degree on different graph datasets share some similar properties, \textit{e.g.}, the power law of degree distribution with different exponent values~\cite{albert2002statistical,qiu2020gcc}. Last but not least, degree-based one-hot embedding is computationally friendly, which avoids high computational costs for embedding initialization. 

In the meantime, to capture global structural knowledge, we introduce a random walk-based structure embedding (RWSE) which is computed based on the random walk diffusion process~\cite{tong2006fast,dwivedi2021graph,liu2023beyond}. Concretely, RWSE is denoted as 
\begin{equation}
\mathbf{s}_{v}^{\mathrm{RWSE}}=\left[{\mathbf{T}}_{i i}, {\mathbf{T}}_{i i}^{2}, \cdots, {\mathbf{T}}_{i i}^{k_2}\right] \in \mathbb{R} ^{\mathrm{k_2}}
\end{equation}
\noindent where ${\mathbf{T}}=\mathbf{A}\mathbf{D}^{-1}$ is a random walk transition matrix computed by binary adjacency matrix $\mathbf{A}$ and diagonal degree matrix $\mathbf{D}$, $i$ is the index of node $v$, and $k_2$ is the dimension of RWSE, denoting that neighbors within $k_2$ hops are involved during the diffusion process and contribute to the structure representation $\mathbf{s}_{v}^{\mathrm{RWSE}}$~\cite{dwivedi2021graph}. {The $k$-th element} of $\mathbf{s}_{v}^{\mathrm{RWSE}}$ refers to the landing probability of node $v$ to itself at a $k$-step random walk. Different from DSE that focuses on local geometric property, {RWSE characterizes the unique role (\textit{e.g.}, tail node or central node) of nodes from a global perspective.} Moreover, RWSE is capable of incorporating domain-invariant structure information, such as the distribution of neighborhoods at different hops. 

Finally, the structure embedding can be obtained by concatenating both DSE and RWSE and represented as
\begin{equation}
\mathbf{s}_{v}=\mathrm{concat}[\mathbf{s}_{v}^{\mathrm{DSE}}, \mathbf{s}_{v}^{\mathrm{RWSE}}].
\end{equation}

It is worth noting that \ourmethod is agnostic to the construction of structure embedding, indicating that \ourmethod can be equipped with more types of structure-related embeddings, such as Laplacian eigenvectors~\cite{dwivedi2021graph} and distance-based embeddings~\cite{you2019position}]. We leave this exploration for future work. 

\subsection{Feature-Structure Decoupled GNN}

In most existing federated graph learning (FGL) frameworks~\cite{xie2021federated,zhang2021subgraph}, the local encoders are usually defined as single-channel GNNs, such as GraphSAGE~\cite{hamilton2017inductive} and GIN~\cite{xu2019powerful}, where the only pathway to generate node representations starts from the raw features. However, in FGL, especially the inter-graph task over non-IID graphs, the local datasets of clients are usually from different domains~\textit{e.g.}, molecules, protein, and social networks~\cite{Morris+2020}, resulting in their diverse feature distributions. In this case, if we directly build and share the feature-based encoders across heterogeneous clients, the personalized performance would be seriously hurt due to the {misaligned} feature/representation spaces. 
Moreover, although we have already obtained the structure embeddings containing shareable structure information, how to further extract universal high-level structural knowledge apart from domain-specific features still remains challenging for single-channel GNNs. 

To address the above limitations, inspired by~\cite{dwivedi2021graph}, we propose a feature-structure decoupled GNN that learns attributive and structural knowledge separately with two parallel channels. 
In the feature-based channel, the feature encoder learns from raw node features and generates hidden embeddings by aggregating neighboring attributive and structural information in each layer. Meanwhile, in the structure-based channel, the structure encoder learns on top of structure embeddings and propagates structure information over the graph. 
An overview of the proposed GNN is illustrated in Fig.~\ref{fig:local_gnn}. The GNN model is composed of three parts, (i)~$f_h^{(0)}$ and $f_g^{(0)}$: linear layers with learnable parameters $w_h^{(0)}$ and $w_g^{(0)}$; (ii)~$f_h^{(l)}$ and $f_g^{(l)}$, $l \in \{1,\cdots,L\}$: $L$ stacked feature-structure decoupled GNN layers with each layer parameterized by $w_h^{(l)}$ and $w_g^{(l)}$; (iii)~$f_h^{(C)}$: the classifier parameterized by $w_h^{(C)}$.

Given an input graph $G=(V,E)$, we take a node $v\in V$ with feature vector $\mathbf{x}_v$ and structure embedding $\mathbf{s}_v$ as an example. At the first step, we transform $\mathbf{x}_v$ and $\mathbf{s}_v$ into the corresponding input embeddings $\mathbf{h}_v^{(0)}$ and $\mathbf{g}_v^{(0)}$ by two parallel linear layers $f_h^{(0)}$ and $f_g^{(0)}$, respectively. With such linear layers, the inputs of two channels are unified into vectors with a consistent dimension.

Then, the feature and structure embedding $\mathbf{h}_v^{(0)}$ and $\mathbf{g}_v^{(0)}$ are fed into the $L$-layer GNN. 
At the $l$-th layer of structure encoder, denoted as $f_g^{(l)}$, the output hidden structure embedding $\mathbf{g}_v^{l}$ is generated by aggregating and transforming the hidden structure embeddings output by the previous layer, \textit{i.e.}, $\mathbf{g}_v^{(l-1)}$ and $\{\mathbf{g}_u^{(l-1)}:u\in \mathcal{N}(v)\}$. Note that the structure encoder is only used to learn structural knowledge and hence would not take any signal from features as its input. 
Meanwhile, $f_h^{(l)}$, the $l$-th layer of feature encoder, takes the concatenation of hidden structure embeddings and hidden feature embeddings output by the previous layer as its input, and produces its output hidden feature embedding $\mathbf{h}_v^{(l)}$ with a message propagation-based GNN layer. In this way, the structural knowledge learned by the structure encoder can be further leveraged to generate feature embeddings, providing supplementary information for representation learning. Through our dual-channel GNN, the structure information can be learned independently, benefiting the feature learning procedure. 

After the $L$ stacked layers, all the node-level hidden feature/structure embeddings output by the last GNN layer, denoted by $\{\mathbf{h}_v^{(L)}: v \in V\}$ and $\{\mathbf{g}_v^{(L)}: v \in V\}$, are first concatenated together and then transformed to the graph-level embedding $\mathbf{h}_G$ of graph $G$ with a readout function. In the final step, we optimize the local model with a cross-entropy loss $\mathcal{L}_m$ following the procedures of graph classification tasks. The set of learnable parameters at the $m$-th client is denoted by $w_m=\{w_{h,m},w_{g,m}\}$ where $w_{h,m}$ refers to the parameters in both the feature encoder and the classifier and $w_{g,m}$ refers to those in the structure encoder. They are updated during the local training process as
\begin{equation} \label{eq:local_update}
w_{h,m}^{*}, w_{g,m}^{*} = \underset{w_{h,m},w_{g,m}}{\arg \min} \mathcal{L}_m(w_m; \mathcal{D}_m).
\end{equation}

\noindent \textbf{Discussion:} Compared to conventional single-channel GNNs, the feature-structure decoupled GNN enjoys the following advantages in non-IID FGL scenarios. First, through learning structure embeddings with an independent structure encoder, the structure information of data in various domains can be projected into the same representation space. In this case, the structural knowledge can be universally shared across different domains without the side-effect of unaligned feature space. Moreover, the feature encoder can focus on learning the domain-specific knowledge from the local feature space, and further benefit from structural knowledge. 

\subsection{Structural Knowledge Sharing}
Based on the universal structure information captured by structure embeddings and structure encoders, \ourmethod aims to share the learned structural knowledge across clients with data from diverse domains. Fig.~\ref{fig:agg_scheme} illustrates the pipeline of our learning paradigm, where clients own graph datasets from different domains and the same model architecture. For the $m$-th client, the local model with parameters $w_m$ can be explicitly decoupled as two submodels, \textit{i.e.}, the feature encoder with parameters $w_{h,m}$ and the structure encoder with parameters $w_{g,m}$. {Our core idea is to share $w_{g,m}$ with the FL framework while keeping $w_{h,m}$ being trained locally.} 

The overall training procedure of \ourmethod is similar to FedAvg~\cite{mcmahan2016communication} where the local training at the client-side and central aggregation at the server are iteratively conducted. First, clients transmit their structure encoder parameters $\{w_{g,m}\}|_{m=1}^M$ to the server. 
Next, the server performs a weighted average of these local parameters to obtain the global structure encoder $\overline{w}_g$,
\begin{equation} \label{eq:wg_agg}
\overline{w}_g = \sum_{m=1}^{M} \frac{\left|\mathcal{D}_{m}\right|}{N} {w}_{g,m},
\end{equation}
where $|\mathcal{D}_{m}|$ refers to the number of graphs in the local dataset of client $m$ and $N$ refers to the total number of graphs across all clients. Then, the server returns $\overline{w}_g$ to clients, and clients update their local structure encoder by $\overline{w}_g$ and start local training for the next round.

\begin{table*}[htbp!] 
    \small
    \caption{Performance on different federated graph classification tasks. In each task/setting, there are multiple datasets owned by different clients. These datasets come from one or multiple domains.}
    \vspace{-0.2cm}
    \label{tab:perf}
    \centering
    \begin{tabular}{c cc cc cc cc}
	\toprule
	\textbf{Setting (\# domains)} & \multicolumn{2}{c}{CHEM(1)} & \multicolumn{2}{c}{BIO-CHEM(2)} & \multicolumn{2}{c}{BIO-CHEM-SN(3)} & \multicolumn{2}{c}{BIO-SN-CV(3)} \\
	\cmidrule(lr){1-1} \cmidrule(lr){2-3} \cmidrule(lr){4-5} \cmidrule(lr){6-7} \cmidrule(lr){8-9}
	\textbf{\# datasets} & \multicolumn{2}{c}{7} & \multicolumn{2}{c}{10} & \multicolumn{2}{c}{13} & \multicolumn{2}{c}{9} \\
	\cmidrule(lr){1-1} \cmidrule(lr){2-3} \cmidrule(lr){4-5} \cmidrule(lr){6-7} \cmidrule(lr){8-9}
	\textbf{Accuracy}  &  avg.  &  avg. gain  &  avg.  &  avg. gain  &  avg.  &  avg. gain  &  avg.  &  avg. gain \\
	\midrule
    Local  &  75.38$\pm$2.26  &  -  &  71.09$\pm$1.21  &  -  &  69.37$\pm$3.05  &  -  &  66.91$\pm$2.84  &  -  \\
    \midrule
    FedAvg  &  75.26$\pm$2.00  &  -0.13  &  70.65$\pm$2.73  &  -0.44  &  68.92$\pm$2.12  &  -0.45  &  64.86$\pm$2.73  &  -2.05  \\
    FedProx  &  75.30$\pm$2.00  &  -0.08  &  70.75$\pm$2.26  &  -0.34  &  69.21$\pm$2.63  &  -0.16  &  65.18$\pm$2.01  &  -1.72  \\
    FedPer  &  77.09$\pm$3.36  &  1.70  &  71.97$\pm$1.97  &  0.88  &  69.37$\pm$2.92  &  -0.01  &  62.23$\pm$3.76  &  -4.67  \\
    \midrule
    FedSage  &  75.90$\pm$1.85  &  0.51  &  70.34$\pm$1.87  &  -0.74  &  69.55$\pm$2.15  &  0.18  &  67.95$\pm$1.87  &  1.04  \\
    GCFL  &  76.49$\pm$1.23  &  1.11  &  71.60$\pm$2.20  &  0.51  &  70.65$\pm$1.84  &  1.28  &  66.31$\pm$2.36  &  -0.60  \\
    \ourmethod (Ours)  &  \textbf{79.79}$\pm$2.44  &  \textbf{4.41}  &  \textbf{74.54}$\pm$2.50  &  \textbf{3.46}  &  \textbf{72.16}$\pm$2.43  &  \textbf{2.78}  &  \textbf{69.49}$\pm$1.81  &  \textbf{2.58}  \\
    \bottomrule
    \end{tabular}
    \vspace{-0.2cm}
\end{table*}

Such a structural knowledge sharing scheme helps clients to build a generalized structure encoder. Since the structural encoder is purely based on the structural information without any feature information, it is prone to capture more domain-invariant patterns behind the graphs. Then, with the aggregation at the server, \ourmethod tends to extract the common knowledge maintained by these structure encoders. Through iterative training and aggregating the structure-based encoder, the structural knowledge is generalized over non-IID graphs from various domains. 
Meanwhile, globally shared structural knowledge can further benefit the specific learning tasks at each client. 
With the shortcut that connects structure-based embedding to feature-based embedding at each layer, the underlying globally shared structural knowledge further guides the feature learning process as additional structure-aware information. To sum up, with our proposed structural knowledge sharing scheme, the FGL framework is capable of exploring the common structural knowledge as well as boosting the representation ability of domain-specific feature learning.

\section{Experiments} \label{sec:experiments}
\subsection{Experimental Setup} 
\paragraph{Datasets.}
Following the settings in~\cite{xie2021federated}, we use 16 public graph classification datasets from four different domains, including Small Molecules (MUTAG, BZR, COX2, DHFR, PTC\_MR, AIDS, NCI1), Bioinformatics (ENZYMES, DD, PROTEINS), Social Networks (COLLAB, IMDB-BINARY, IMDB-MULTI), and Computer Vision (Letter-low, Letter-high, Letter-med)~\cite{Morris+2020}. To simulate the data heterogeneity in FGL, we create four non-IID settings, \textit{i.e.}, (1) cross-dataset setting based on seven small molecules datasets (CHEM); (2)-(4) cross-domain setting based on datasets from two or three domains (BIO-CHEM, BIO-CHEM-SN, BIO-SN-CV). In each of the settings, a client owns one of the corresponding datasets and randomly splits it into three parts: 80\% for training, 10\% for validation, and 10\% for testing. 

\paragraph{Baselines.} 
We compare \ourmethod with six baselines including (1) \textbf{Local} where clients train their model locally;  (2)~\textbf{FedAvg}~\cite{mcmahan2016communication}, the standard FL algorithm; (3)~\textbf{FedProx}~\cite{li2018federated} and (4)~\textbf{FedPer}~\cite{arivazhagan2019federated} that deal with heterogeneity issues in FL; (5)~\textbf{FedSage}~\cite{zhang2021subgraph} and (6)~\textbf{GCFL}~\cite{xie2021federated}, two state-of-the-art FGL methods.

\paragraph{Implementation Details.}
We use a three-layer GCN~\cite{kipf2017semi} as the structure encoder and a three-layer GIN~\cite{xu2019powerful} as the feature encoder, both with the hidden size of 64. The dimension of DSE and RWSE, denoted as $k_1$ and $k_2$, are both set to be 16. The local epoch number and batch size are 1 and 128, respectively. We use an Adam~\cite{kingma2014adam} optimizer with weight decay 5e-4 and learning rate 0.001. The number of communication rounds is 200 for all FL methods. We report the results with the average over 5 runs of different random seeds. We implement all the methods using PyTorch and conduct all experiments on one NVIDIA Tesla V100 GPU. More implementation details about the model architecture, datasets, hyper-parameters, and baselines can be found in Appendix~A. The code of \ourmethod is available at \url{https://github.com/yuetan031/FedStar}.

\subsection{Experimental Results}
\paragraph{Performance Comparison.} 
We show the federated graph classification results of all methods under four non-IID settings, including one cross-dataset setting (CHEM) and three cross-domain settings (BIO-CHEM, BIO-CHEM-SN, BIO-SN-CV). We summarize the final average test accuracy and its average gain compared with Local in Table~\ref{tab:perf}.
It suggests that \ourmethod outperforms all the baselines by a notable margin. 
Conventional FL methods such as FedAvg and FedProx fail to surpass Local due to their inevitable performance deterioration in non-IID settings, while personalized FL algorithm FedPer achieves better performance because only partial learnable parameters are aggregated at the server, alleviating the deterioration issue. Two FGL methods, \textit{i.e.}, FedSage and GCFL, perform better than Local in most cases, owing to their unique designs for FGL tasks. For example, GCFL introduces the clustering scheme to aggregate within similar clients. 

\begin{figure*}[t!] 
  \setlength{\abovecaptionskip}{-0.1cm}
  \setlength{\belowcaptionskip}{-0.2cm}
  \caption{Test accuracy curves of our proposed \ourmethod and five FL/FGL methods along the communication rounds. Each subfigure corresponds to a specific cross-dataset (a) or cross-domain (b,c,d) non-IID settings. }
  \label{fig:curves}
  \begin{center}
    \subfigure[CHEM]{
		\begin{minipage}[b]{0.23\textwidth}
			\includegraphics[width=1\textwidth]{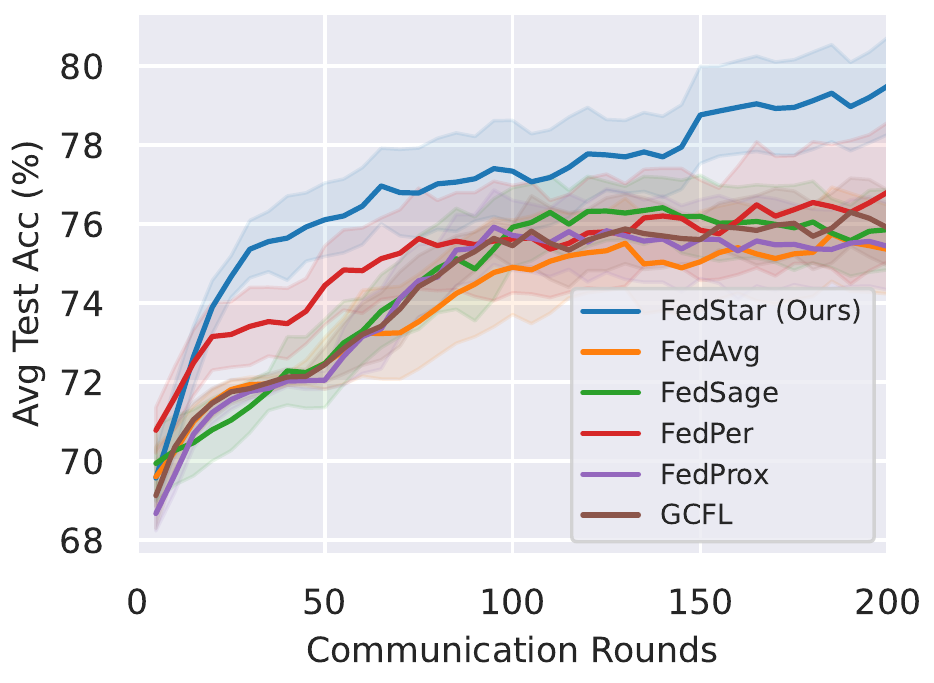}
		\end{minipage}
		\label{fig:curve_chem}
	}	
	\subfigure[BIO-CHEM]{
		\begin{minipage}[b]{0.23\textwidth}
			\includegraphics[width=1\textwidth]{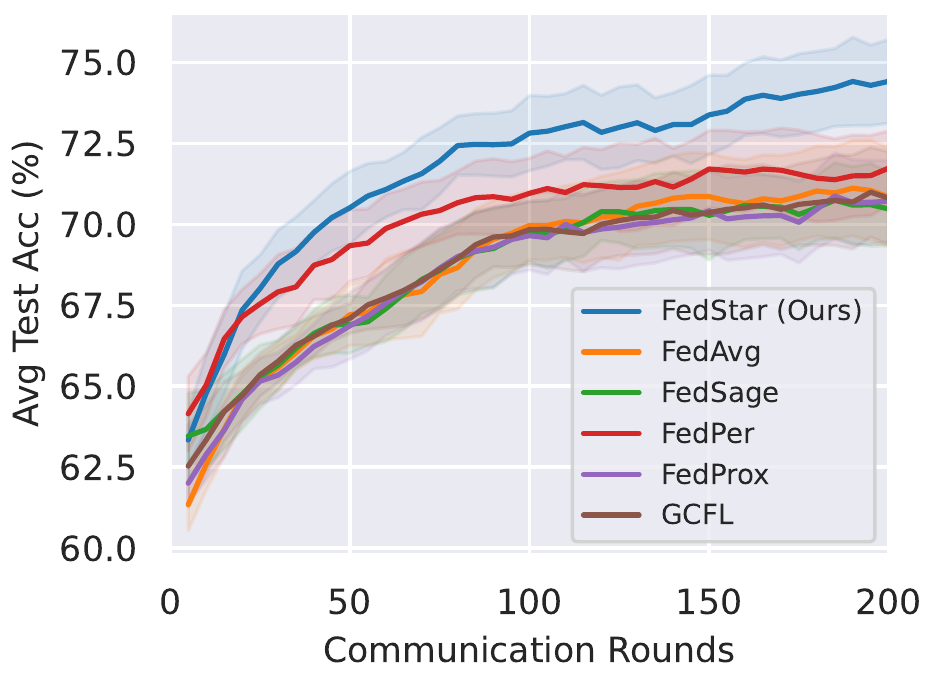}
		\end{minipage}
		\label{fig:curve_biochem}
	}
	\subfigure[BIO-CHEM-SN]{
		\begin{minipage}[b]{0.23\textwidth}
			\includegraphics[width=1\textwidth]{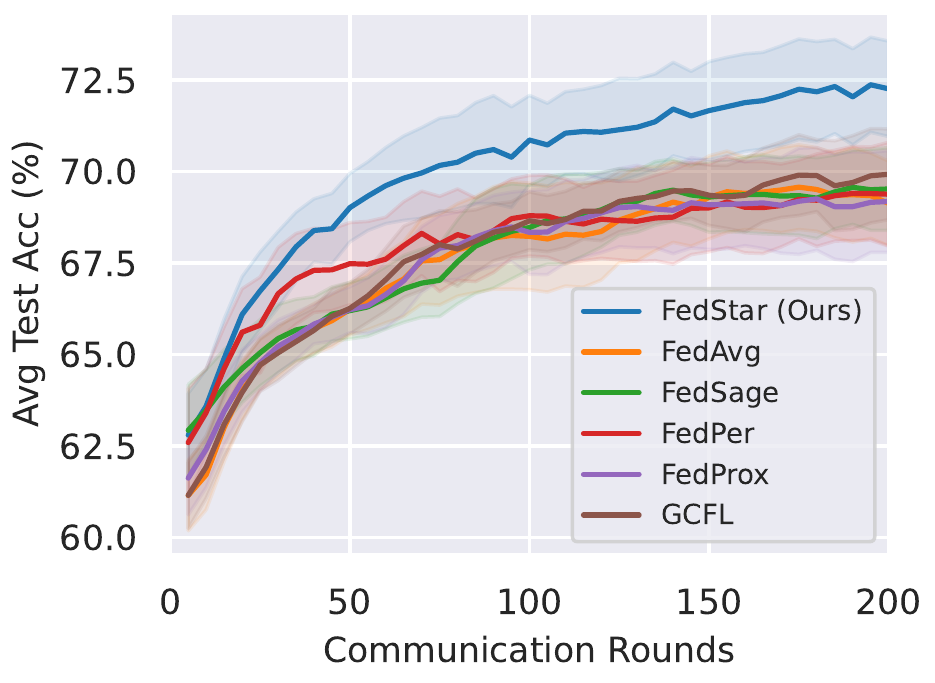}
		\end{minipage}
		\label{fig:curve_biochemsn}
	}
	\subfigure[BIO-SN-CV]{
		\begin{minipage}[b]{0.23\textwidth}
			\includegraphics[width=1\textwidth]{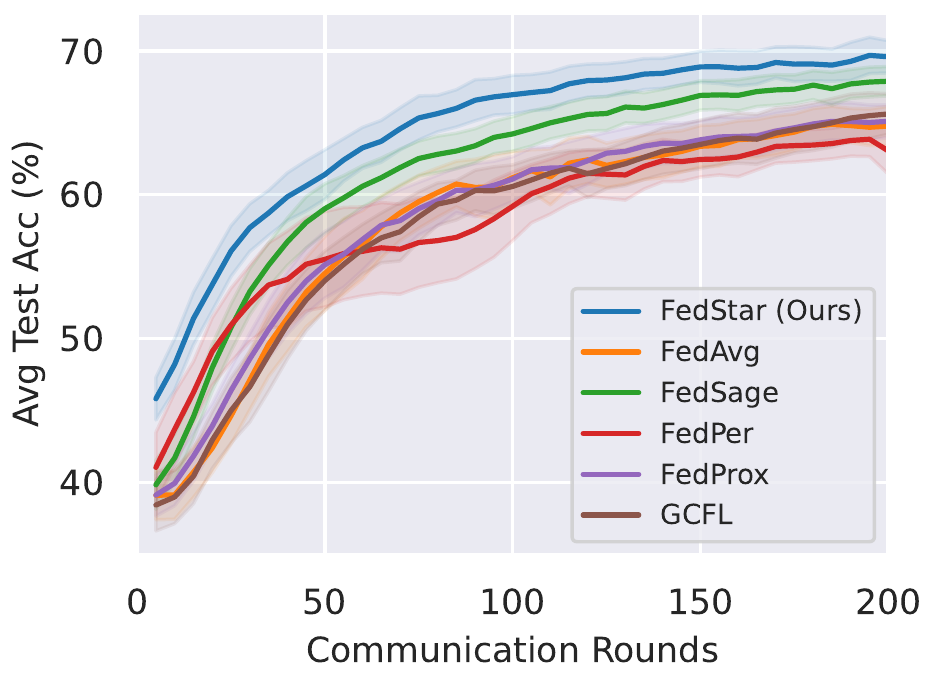}
		\end{minipage}
		\label{fig:curve_biosncv}
	}
	\vspace{-0.2cm}
  \end{center}
  \vspace{-0.4cm}
\end{figure*}

\paragraph{Convergence Analysis.}
Fig.~\ref{fig:curves} shows the curves of the average test accuracy with standard deviation during the training process across five random runs, including the results of all the baselines. It can be observed that in all the four non-IID settings, \ourmethod dominates the other methods on the average test accuracy and achieves a faster convergence, suggesting that it is much easier to align the structure-based encoder compared with aligning the feature-based encoder.

\begin{table}[t]
\centering
\caption{Ablation studies on the effects of decoupling and sharing mechanisms. FE and SE stand for feature encoder and structure encoder, respectively. FE: feature encoder; All: all learnable parameters; DC: decoupling.} 
\label{tab:abl_share_decouple}
\vspace{-0.3cm}
    \resizebox{1\columnwidth}{!}{
    \begin{tabular}{ccccc}
    \toprule
    
    \multirow{2}{*}{\textbf{Sharing}} & \multirow{2}{*}{\textbf{DC}} & \multicolumn{3}{c}{\textbf{Setting (\# domains)}} \\
    \cmidrule{3-5}
      &  & BIO-CHEM(2) & BIO-CHEM-SN(3) & BIO-SN-CV(3) \\ 
    \midrule
    All & - & 70.86$\pm$2.25 & 69.32$\pm$2.42 & 65.23$\pm$2.52 \\ 
    None & - & 71.59$\pm$1.93 & 69.42$\pm$3.06 & 68.17$\pm$3.04 \\ 
    All & \checkmark & 71.97$\pm$2.14 & 69.85$\pm$2.43 & 65.78$\pm$4.25 \\ 
    None & \checkmark & 74.08$\pm$2.45 & 71.30$\pm$1.89 & 68.76$\pm$2.24 \\ 
    FE & \checkmark & 71.00$\pm$3.51 & 68.53$\pm$2.74 & 64.14$\pm$2.73 \\
    \midrule
    SE(Ours) & \checkmark & \textbf{74.54$\pm$2.50} & \textbf{72.15$\pm$2.43} & \textbf{69.49$\pm$1.81} \\ 
    \bottomrule
    \end{tabular}
    }
\vspace{-0.3cm}
\end{table}

\begin{table}[ht]
\vspace{-0.2cm}
\centering
\caption{An ablation study on varying structure embeddings.}
\label{tab:abl_se}
\vspace{-0.3cm}
    \resizebox{1\columnwidth}{!}{
    \begin{tabular}{ccccc}
    \toprule
    \multirow{2}{*}{\textbf{DSE}} & \multirow{2}{*}{\textbf{RWSE}} & \multicolumn{3}{c}{\textbf{Setting (\# domains)}} \\
    \cmidrule{3-5}
      &   & BIO-CHEM(2) & BIO-CHEM-SN(3) & BIO-SN-CV(3) \\ 
    \midrule
    - & - & 69.51$\pm$2.25 & 69.64$\pm$1.92 & 66.05$\pm$2.92 \\
    \checkmark & - & 74.42$\pm$3.15 & 72.05$\pm$2.82 & 69.25$\pm$2.41 \\
    - & \checkmark &  72.74$\pm$3.44 & 70.48$\pm$3.37 & 67.23$\pm$2.74 \\
    \checkmark & \checkmark & \textbf{74.54$\pm$2.50} & \textbf{72.15$\pm$2.43} & \textbf{69.49$\pm$1.81} \\
    \bottomrule
    \end{tabular}
    }
\vspace{-0.0cm}
\end{table}

\begin{table}[ht]
\centering
\caption{The performance on cross-dataset non-IID setting (CHEM), with varying local epochs.}
\label{tab:varying_e}
\vspace{-0.3cm}
    \resizebox{1\columnwidth}{!}{
    \begin{tabular}{ccccc}
    \toprule
    Epochs & 1 & 2 & 3 & 4 \\
    \midrule
    FedAvg &   75.26$\pm$2.00 & \textbf{76.59$\pm$2.72} & 76.96$\pm$3.30 & 76.36$\pm$3.36 \\
    GCFL & \textbf{76.49$\pm$1.23} & 76.23$\pm$2.99 & 75.88$\pm$2.65 & 76.27$\pm$2.14 \\
    FedStar & 79.79$\pm$2.44 & 80.26$\pm$1.89 & 79.90$\pm$2.96 & \textbf{80.58$\pm$2.48} \\
    \bottomrule
    \end{tabular}
    }
\vspace{-0.4cm}
\end{table}

\paragraph{Effects of Decoupling and Sharing Mechanisms.}
We compare several variants of \ourmethod by sharing different components of the local GNN (\textit{i.e.}, all parameters, feature encoder, structure encoder, and none) and/or involving feature-structure decoupled scheme or not. For the variants without decoupling scheme, we alternatively remove the structure encoder and concatenate the structure embedding with the raw feature for once. 
In Table~\ref{tab:abl_share_decouple}, we observe that the feature-structure decoupling scheme brings about 0.5\%-2.4\% improvement to the cases without it, indicating that independently learning structural and attributive knowledge can improve FGL performance in general. 
Moreover, compared with pure local training, sharing all parameters or the feature encoder results in performance degradation. Such an observation illustrates that sharing feature information across heterogeneous datasets perturbs the FGL procedure, supporting our interpretation in Fig.~\ref{fig:intro_deter}. Finally, we find that only sharing the structure encoder leads to the best result, validating the effectiveness of the structural knowledge sharing paradigm. 

\vspace{-0.2cm}
\paragraph{Effects of Different Structure Embeddings.}
To understand how the explicit structure embedding contributes to the final performance, we carry out an ablation study by varying the components of structure embeddings. For the variant without DSE and RWSE, we take all-ones vectors as the structure embedding. 
As shown in Table~\ref{tab:abl_se}, by concatenating DSE and RWSE, the best performance is achieved in all the three non-IID settings. It also suggests that, DSE plays a more important role than RWSE when considering only one structure embedding, which means that local structural information is more prone to be globally shared by different domains than global structural information. 
Moreover, the contribution made by RWSE is still not negligible and can benefit the structure learning. 

\paragraph{Varying Local Epochs.}
In FL, clients can perform multiple local training epochs before global aggregation to reduce the communication costs. In Table~\ref{tab:varying_e}, we provide the results of FedAvg, GCFL, and \ourmethod on the non-IID setting CHEM where there are seven clients in total, each of which owns a unique small molecule dataset. The results show that \ourmethod benefits more from the increasing local epochs and thus more communication-efficient.

\begin{figure}[t!]
\vspace{-0.2cm}
	\centering
	\setlength{\abovecaptionskip}{-0.0cm}
	\setlength{\belowcaptionskip}{-0.2cm}
	\includegraphics[width=0.8\columnwidth]{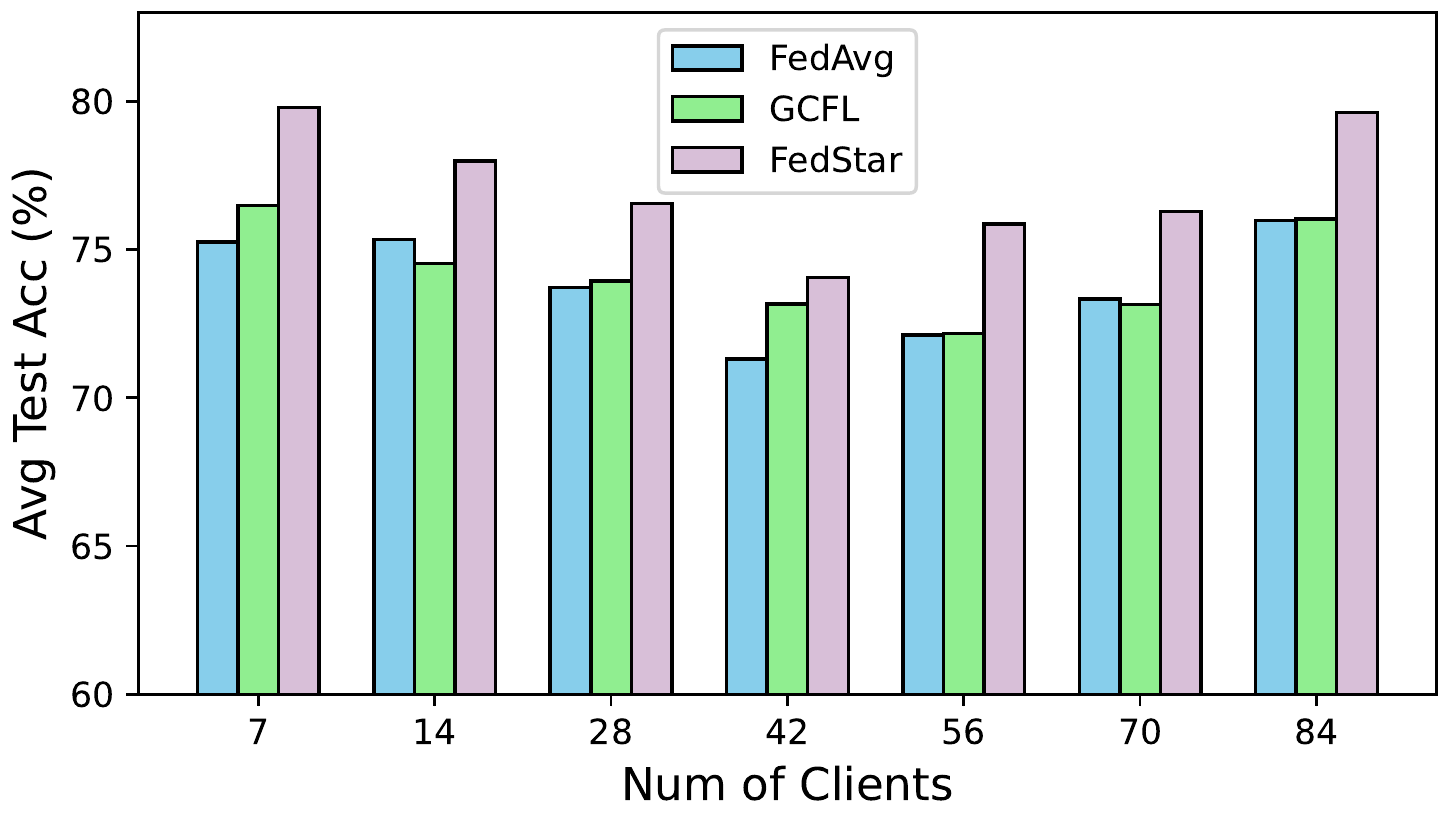}
	\caption{The performance on cross-dataset non-IID setting (CHEM), with varying numbers of clients.}
	\label{fig:varying_client}
	\vspace{-0.2cm}
\end{figure}

\vspace{-0.1cm}
\paragraph{Varying Numbers of Clients.}
To test the performance of \ourmethod in a larger FGL framework, we increase the number of clients from 7 to 84 under the CHEM non-IID setting where each of the seven small molecules datasets are split to 1-12 shards and then each shard is assigned to one client. 
In Fig.~\ref{fig:varying_client}, as the number of clients increases, the performance of all the three algorithms first drops because of the further divergence when there are more clients participating training, and then improves because the local task in each client becomes easier when the local dataset is small enough. It suggests that \ourmethod outperforms the other two algorithms in both small- and large-scale FGL systems.

\vspace{-0.1cm}
\section{Conclusion} \label{sec:conclusion}
In this paper, we propose a novel federated graph learning (FGL) framework, namely \ourmethod, that solves the non-IID issues via structural knowledge sharing. 
Based on structure embeddings and feature-structure decoupled GNN, structural knowledge is explicitly represented and well captured without being influenced by the feature-based information. 
Through sharing the structure encoder within the FL framework, our proposed framework enables clients from different domains to learn domain-agnostic structural knowledge globally while keeping their feature-based node representation learning personalized. 
Experimental results illustrate that \ourmethod consistently outperforms the state-of-the-art methods in various non-IID FGL scenarios.

\clearpage
\bibliography{aaai23}

\clearpage
\appendix
\onecolumn
\section{Appendix}

\subsection{A Experimental Details}
We provide more experimental details here due to the page limit.

\subsubsection{A.1 Detailed Experimental Setup}
We select some important hyper-parameters through small grid search on the validation dataset, and keep the rest insensitive hyper-parameters to be fixed values. Concretely, the grid search is carried out on the following search space:
\begin{itemize}
    \item The dimension of degree-based structure embedding $k_1$: \{8,16,32,64,128\}
    \item The dimension of random walk-based structure embedding $k_2$: \{8,16,32,64,128\}
    \item The learning rate: \{5e-3, 1e-3, 5e-4, 1e-4\}
    \item The weight decay: \{7e-4,5e-4,3e-4,1e-4\}
\end{itemize}

\subsubsection{A.2 Details of the Baseline Methods}
We compare FedStar with six baselines. The details of these baselines are provided as follows.
\begin{itemize}
    \item \textbf{Local}: Each client trains their local model based on the local data without communication with others. 
    \item \textbf{FedAvg}: Clients send all the learnable parameters to the server and receive the aggregated parameters from the server for their next-round training.
    \item \textbf{FedProx}: Based on FedAvg, a regularization term with importance weight $\mu$ is added to the original loss function. In our experiments, $\mu$ is set to 0.01.
    \item \textbf{FedPer}: Compared with FedAvg, instead of sharing all the learnable parameters, only the parameters in graph convolution layers are shared across clients.
    \item \textbf{FedSage}: A three-layer GraphSage model is trained under the FedAvg framework. 
    \item \textbf{GCFL}: Compared with FedAvg, instead of sharing parameters within all participating clients, sharing is conducted in a small group/cluster consisting of similar clients. There are two hyper-parameters determining the clustering results, \textit{i.e.}, $\epsilon_1$ and $\epsilon_2$. We use the values in~\cite{xie2021federated} to guarantee the performance of GCFL.
\end{itemize}

\subsubsection{A.3 Data Splitting Details}
As we mentioned in the main paper, there are four non-IID settings including one cross-dataset setting, namely CHEM, and three cross-domain settings, namely BIO-CHEM, BIO-CHEM-SN, and BIO-SN-CV. Among these non-IID settings, CHEM, BIO-CHEM, and BIO-CHEM-SN follow from the empirical studies in~\cite{xie2021federated}, BIO-SN-CV is first proposed in this work. All the datasets are graph classification datasets and publicly available in~\cite{Morris+2020}. 

The non-IID data splitting details of Table~\ref{tab:perf} are given as follows.

\begin{table*}[htbp!]
    \caption{Detailed statistics of \textbf{CHEM} in all the non-IID settings in Table~\ref{tab:perf}.}
	\label{tab:chem_data}
    \centering
    \begin{tabular}{c|ccccccc}
	\toprule
	Domain & \multicolumn{7}{c}{Small Molecules} \\
	\midrule
	Datasets & MUTAG & BZR & COX2 & DHFR & PTC\_MR & AIDS & NCI1 \\
	\midrule
	\# of clients & 1 & 1 & 1 & 1 & 1 & 1 & 1 \\
	\# of graphs for training per client & 150 & 324 & 373 & 604 & 275 & 1600 & 3288 \\
	\# of nodes per graph  & 17.95 & 35.87 & 41.14 & 42.43 & 14.19 & 15.75 & 29.98 \\
	\bottomrule
    \end{tabular}
\end{table*}

\begin{table*}[htbp!]
	\caption{Detailed statistics of \textbf{BIO} in all the non-IID settings in Table~\ref{tab:perf}.}
	\label{tab:bio_data}
    \centering
    \begin{tabular}{c|ccc}
	\toprule
	Domain & \multicolumn{3}{c}{Bioinformatics} \\
	\midrule
	Datasets & ENZYMES & DD & PROTEINS \\
	\midrule
	\# of clients & 1 & 1 & 1\\
	\# of graphs for training per client & 480 & 942 & 890 \\
	\# of nodes per graph & 33.19 & 280.94 & 38.07 \\
	\bottomrule
    \end{tabular}
\end{table*}

\begin{table*}[htbp!]
	\caption{Detailed statistics of \textbf{SN} in all the non-IID settings in Table~\ref{tab:perf}.}
	\label{tab:sn_data}
    \centering
    \begin{tabular}{c|ccc}
	\toprule
	Domain & \multicolumn{3}{c}{Social Networks} \\
	\midrule
	Datasets & COLLAB & IMDB-BINARY & IMDB-MULTI \\
	\midrule
	\# of clients & 1 & 1 & 1\\
	\# of graphs for training per client & 4000 & 800 & 1200 \\
	\# of nodes per graph & 73.92 & 19.84 & 13.03 \\
	\bottomrule
    \end{tabular}
\end{table*}

\begin{table*}[htbp!]
	\caption{Detailed statistics of \textbf{CV} in all the non-IID settings in Table~\ref{tab:perf}.}
	\label{tab:cv_data}
    \centering
    \begin{tabular}{c|ccc}
	\toprule
	Domain & \multicolumn{3}{c}{Computer Vision} \\
	\midrule
	Datasets & Letter-low & Letter-high & Letter-med \\
	\midrule
	\# of clients & 1 & 1 & 1\\
	\# of graphs for training per client & 1800 & 1800 & 1800 \\
	\# of nodes per graph & 4.65 & 4.68 & 4.70 \\
	\bottomrule
    \end{tabular}
\end{table*}

\newpage
\subsubsection{A.4 Model Architecture}
We use a three-layer GCN as the structure encoder and a three-layer GIN as the feature encoder. The concrete model architecture is shown in Table~\ref{tab:model_setting}. 

\begin{table*}[htbp]
	\caption{The model architecture at each client. There are eight layers in total. The components shared across clients in our method are in \textbf{BOLD}. FC: fully connected layer. n\_feat and n\_class: the dimension of node features and the number of graph classes at a specific client.}
	\label{tab:model_setting}
    \centering
    \begin{tabular}{c|ll}
	\toprule
	Layer & \multicolumn{2}{c}{Details} \\
	\midrule
	1 & FC(n\_feat, 64) & \textbf{FC(32, 64)} \\
	\midrule
	2 & GIN(128,64), ReLU, Dropout(0.5) & \textbf{GCN(64,64)}, Tanh \\
	\midrule
	3 & GIN(128,64), ReLU, Dropout(0.5) & \textbf{GCN(64,64)}, Tanh \\
	\midrule
	4 & GIN(128,64), ReLU, Dropout(0.5) & \textbf{GCN(64,64)}, Tanh \\
	\midrule
	5 & POOLING & \\
	\midrule
	6 & FC(128,64) & \\
	\midrule
	7 & FC(64,64), ReLU, Dropout(0.5) & \\
	\midrule
	8 & FC(64,n\_class) & \\
	\bottomrule
    \end{tabular}
\end{table*}

\newpage
\subsection{B Addition Experimental Results}
We conduct experiments on CHEM setting using various GNN models. In Table~\ref{tab:various_gnn_backbone}, we show that FedStar has stable and considerable performance with various GNNs as its feature (F) and structure (S) encoder, verifying its universality on different GNN backbones.

\begin{table*}[h]
\centering
\caption{The performance of FedStar when using various GNN backbones as feature encoder (FE) and structure encoder (SE).} 
\label{tab:various_gnn_backbone}
    \begin{tabular}{cccc}
    \toprule
    FE/SE & GCN & GIN & SAGE \\
    \midrule
    GCN & 76.96 & 77.15 & 78.41 \\ 
    GIN & 79.79 & 77.63 & 79.09 \\ 
    SAGE & 78.89 & 79.04 & 78.86 \\ 
    \bottomrule
    \end{tabular}
\end{table*}

\subsection{C Algorithm}

\begin{algorithm*}[htbp!]
	\caption{{\bf \ourmethod} (\textbf{Fed}erated Graph Learning via \textbf{St}ructural Knowledge Sh\textbf{ar}ing)} 
	\hspace*{0.02in} {\bf Server input:} 
	initial structure GNN encoder $\overline{w}_g$; \\
	\hspace*{0.02in} {\bf Client $m$'s input:} initial feature GNN encoder $w_{h,m}$, local labeled graph dataset $\mathcal{D}_m$;\\
	\hspace*{0.02in} {\bf Server executes:}
	\begin{algorithmic}[1]
	    \STATE Communicate $\overline{w}_g$ to all clients $m \in [1, M]$;
		\FOR{each round $T = 1,2,...$} 
		\FOR{each client $m$ {\bf in parallel}}
		\STATE ${w}_{g,m} \leftarrow$ LocalUpdate $\left(m, \overline{w}_g\right)$;
		\ENDFOR
		\STATE \textbf{Update structure encoder} by $\overline{w}_g = \sum_{m=1}^{M} \frac{\left|\mathcal{D}_{m}\right|}{N} {w}_{g,m}$;
		\ENDFOR
	\end{algorithmic} 
	\hspace*{0.02in} \\
	\hspace*{0.02in} {\bf LocalUpdate}$\left(m, \overline{w}_g\right)$:
	\begin{algorithmic}[1]
	    \STATE Initialize $w_{g,m} \leftarrow \overline{w}_g$;
		\FOR{each local epoch}
		\STATE \textbf{Update local model} by $w_{h,m}^{*}, w_{g,m}^{*} = \underset{w_{h,m},w_{g,m}}{\arg \min} \mathcal{L}_m(w_m; \mathcal{D}_m)$;
		\ENDFOR
		\RETURN $w_{g,m}^{*}$.
	\end{algorithmic}
	\label{alg1}
\end{algorithm*}

\end{document}